\documentclass[acmsmall,screen]{acmart}

\AtBeginDocument{%
  \providecommand\BibTeX{{%
    \normalfont B\kern-0.5em{\scshape i\kern-0.25em b}\kern-0.8em\TeX}}}

\setcopyright{acmcopyright}
\copyrightyear{2022}
\acmYear{2022}
\acmDOI{XXXXXXX.XXXXXXX}

\acmConference[ACM Transactions on Multimedia Computing, Communications, and Applications]{}{}{}
\acmPrice{15.00}
\acmISBN{978-1-4503-XXXX-X/18/06}

\usepackage{subfigure}
\usepackage{multirow}
\usepackage{color}
\usepackage{threeparttable}
\newcommand{\eg}{\emph{e.g.,}~}
\newcommand{\etal}{\emph{et al.}~}
\newcommand{\ie}{\emph{i.e.,}~}

\begin{document}

\title[PLN]{Progressive Localization Networks for Language based Moment Localization}

\author{Qi Zheng}
\email{zhengqi1225@outlook.com}
\affiliation{%
  \institution{Zhejiang Gongshang University}
  \city{Hang Zhou}
  \state{Zhejiang}
  \country{China}
  \postcode{310035}
}

\author{Jiangfeng Dong}
\email{dongjf24@gmail.com}
\affiliation{%
  \institution{Zhejiang Gongshang University}
  \city{Hang Zhou}
  \state{Zhejiang}
  \country{China}
  \postcode{310035}
}

\author{Xiaoye Qu}
\email{xiaoye@hust.edu.cn}
\affiliation{%
  \institution{Huazhong University of Science and Technology}
  \state{Hubei}
  \country{China}
  \postcode{430074}
}

\author{Xun Yang}
\email{xunyang@nus.edu.sg}
\affiliation{%
  \institution{University of Science and Technology of China}
  \country{China}
}

\author{Yabing Wang}
\email{wyb7wyb7@163.com}
\affiliation{%
  \institution{Zhejiang Gongshang University}
  \city{Hang Zhou}
  \state{Zhejiang}
  \country{China}
  \postcode{310035}
}

\author{Pan Zhou}
\email{panzhou@hust.edu.cn}
\affiliation{%
  \institution{Huazhong University of Science and Technology}
  \state{Hubei}
  \country{China}
  \postcode{430074}
}

\author{Baolong Liu}
\email{liubaolongx@gmail.com}
\affiliation{%
  \institution{Zhejiang Gongshang University}
  \city{Hang Zhou}
  \state{Zhejiang}
  \country{China}
  \postcode{310035}
}
\author{Xun Wang}
\email{wx@mail.zjgsu.edu.cn}
\affiliation{%
  \institution{Zhejiang Gongshang University}
  \city{Hang Zhou}
  \state{Zhejiang}
  \country{China}
  \postcode{310035}
}

\renewcommand{\shortauthors}{Q. Zheng et al.}

\begin{abstract}
This paper targets the task of language-based video moment localization. The language-based setting of this task allows for an open set of target activities, resulting in a large variation of the temporal lengths of video moments. Most existing methods prefer to first sample sufficient candidate moments with various temporal lengths, and then match them with the given query to determine the target moment. However, candidate moments generated with a fixed temporal granularity may be suboptimal to handle the large variation in moment lengths. To this end, we propose a novel multi-stage Progressive Localization Network (PLN) which progressively localizes the target moment in a coarse-to-fine manner. Specifically, each stage of PLN has a localization branch, and focuses on candidate moments that are generated with a specific temporal granularity. The temporal granularities of candidate moments are different across the stages. Moreover, we devise a conditional feature manipulation module and an upsampling connection to bridge the multiple localization branches. In this fashion, the later stages are able to absorb the previously learned information, thus facilitating the more fine-grained localization. Extensive experiments on three public datasets demonstrate the effectiveness of our proposed PLN for language-based moment localization, especially for localizing short moments in long videos.
\end{abstract}

\ccsdesc[500]{Information systems~Multimedia and multimodal retrieval}

\keywords{Moment localization, Progressive Learning, Coarse-to-fine Manner, Multi-stage Model.}

\maketitle

\section{Introduction}
\begin{figure}[tb!]
\centering
\subfigure[]{
\label{fig:task_example}
\includegraphics[width=0.45\linewidth]{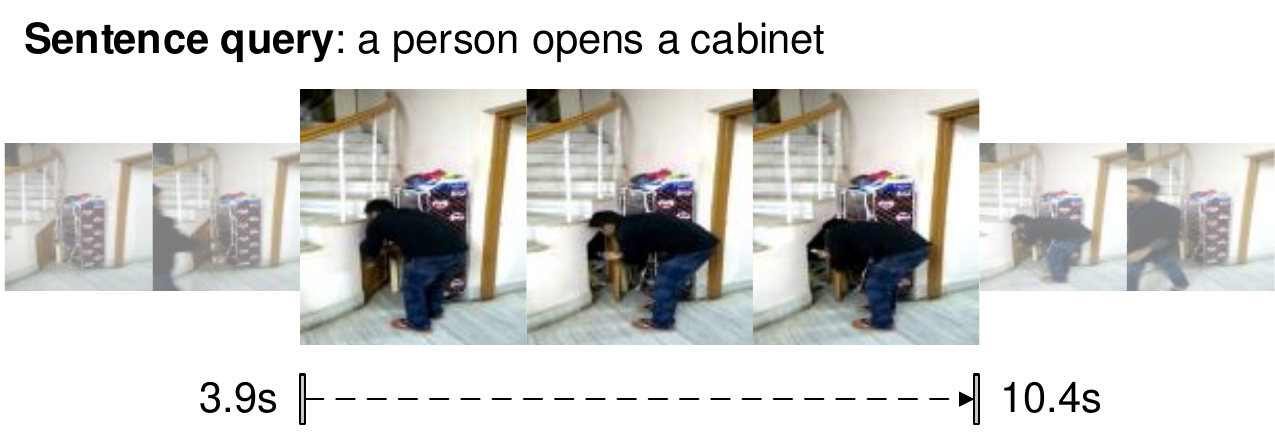}
}
\quad
\subfigure[]{
\includegraphics[width=0.45\linewidth]{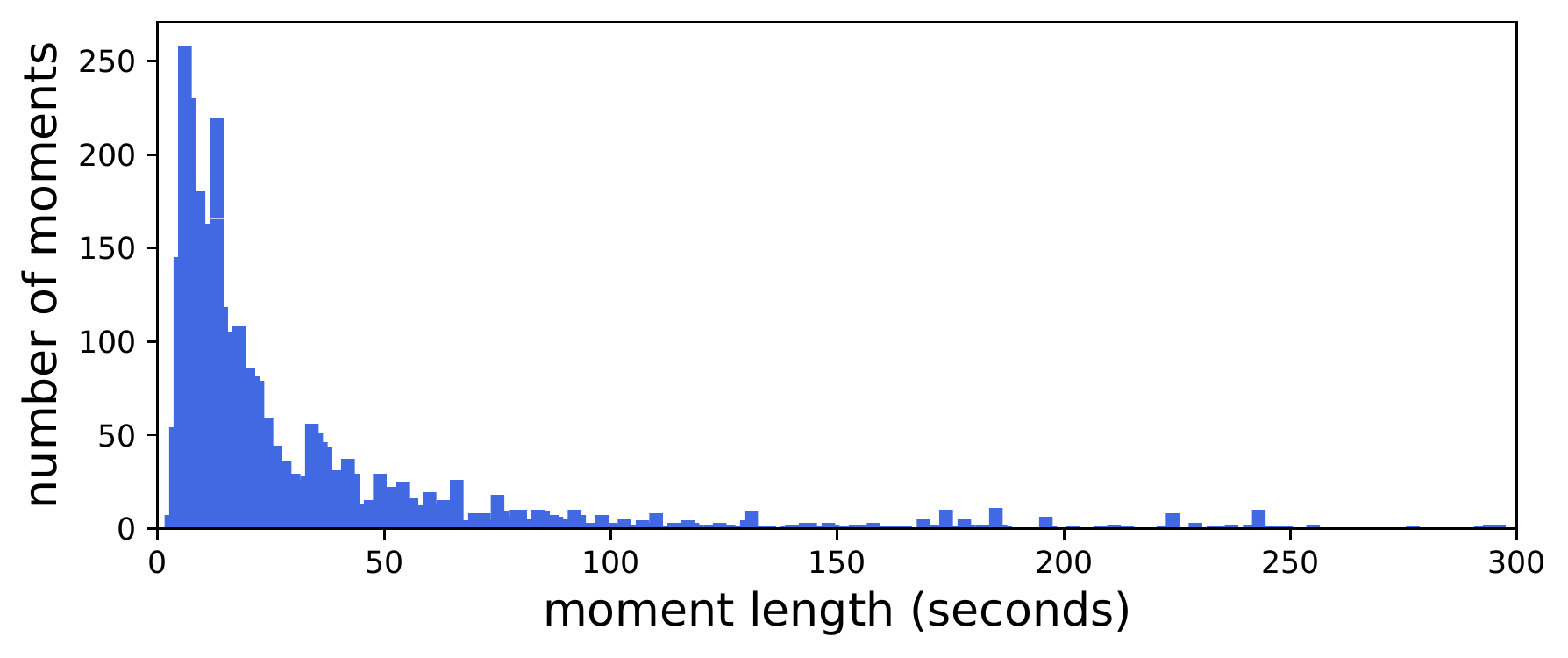}
\label{fig:len_distru}
}
\caption{(a) An example of language-based moment localization. (b) The distribution of target moment lengths on the TACoS dataset, showing a large variation in the temporal lengths of moments.}
 \label{fig:perf_group}
\end{figure}

Localizing actions/activities in videos is an increasingly important but challenging research task for video understanding, which usually can be grouped into two sub-fields: \ie temporal action localization \cite{shou2016temporal,gao2017turn,chen2019relation} and language-based moment localization \cite{yuan2019semantic,liu2021single,gao2021fast}.
For temporal action localization, it aims to temporally localize segments whose action labels are within a pre-defined list of actions. Due to the pre-defined scheme and the limited action labels, it fails to effectively handle unseen activities in the real world.
Therefore, a new task, language-based moment localization, is introduced by Hendricks \etal~\cite{hendricks17iccv} and Gao \etal~\cite{gao2017tall}. 
As shown in Figure \ref{fig:task_example}, given an untrimmed video and a natural language sentence query, the language-based moment localization task aims to temporally localize a specific video segment/moment which semantically matches the given query.
This task is more flexible than temporal action localization, and recently attracts widespread interests in multiple research communities, such as computer vision~\cite{zeng2020dense,mun2020local} and multimedia~\cite{zhang2019mm,cao2020adversarial}.
In this paper, we target the task of language-based moment localization due to its potential applications in intelligent video surveillance, robotics, \textit{etc}.

In the language-based moment localization task, the language-based queries are usually free-form and can describe diverse video content, which allows for an open set of video activities, \eg a short activity of \textit{person closes the door}, or a relatively more complex activity of \textit{the person walks into the house puts down the food}. Such an open setting leads to a large variation in the temporal lengths of the target moments, as shown in Fig. \ref{fig:len_distru}.
Considering such a large variation, how to generate high-quality candidate moments (\textit{a.k.a.} proposals) is one of the key questions to tackle this challenging task.

Most existing works prefer to first sample sufficient candidate moments with various temporal lengths, and then match them with the given query to determine the target moment by their cross-modal similarity
\cite{hendricks17iccv,gao2017tall,Ge_2019_WACV,liu2018attentive,liu2018crossmodal,jiang2019cross,zhang2019learning,liu2020deep,yang2018person}. The popular solution of generating candidate moments mainly consists of two steps. The first step is to split the input video into an ordered sequence of video clips with a fixed interval. The second step is to enumerate~\cite{hendricks17iccv,zhang2019learning} or sample~\cite{hendricks17iccv} contiguous set of clips generated from the first step for obtaining sufficient candidate moments which have various temporal lengths. In this popular strategy, the setting of the split interval for sampling video clips is quite important to the localization performance. A large split interval usually results in a sparse set of video clips where each clip has a relatively large temporal granularity. Such video clips would construct a set of coarse-grained candidate moments that may not be able to effectively handle the target moments of short temporal lengths. In contrast, a small split interval usually results in a dense set of clips of small temporal granularity, which will accordingly generate a set of redundant candidate moments. Such redundant candidates would facilitate the localization of short-length target moments but also inevitably hinder the optimization of the learning objective. 

To alleviate above dilemma, in this paper we explore the video clips of multiple diverse temporal granularities for better handling the moments with large variations in temporal lengths.
We propose to progressively localize the target moment in a coarse-to-fine manner, which is inspired by humans' decision process. When humans perform language-based moment localization, they may first roughly search the target moment and then refine it.
For example, given a query of \textit{a person opens a cabinet} as illustrated in Fig. \ref{fig:task_example}, one may first search the segments where a person appears then further localize the action of opening a cabinet.
To this end, we propose a Progressive Localization Network (PLN), as shown in Fig. \ref{fig:structure}, which progressively localizes the target moment via multiple stages and each stage has a localization branch with a specific temporal granularity. 

Different from existing works \cite{hendricks17iccv,zhang2019learning,liu2018crossmodal} that utilize a fixed interval to produce video clips, we propose to use different intervals across multiple stages resulting in clips of diverse temporal granularities. Specifically, we use a large interval in the early stage and a small interval in the late stage. By this way, the localization branch in the early stage focuses on the moments generated with the clips of large temporal granularity, while the later localization branch focuses more on the moments generated with the clips of a small granularity. Such a coarse-to-fine manner contributes to handling the moments with large variations in temporal lengths.
Moreover, all branches of the model are not independent as we introduce a conditional feature manipulation module and an upsampling connection to bridge them.
Although the model has multiple stages, it can be trained jointly in an end-to-end style.
Our main contributions are roughly summarized as follows:

\begin{itemize}
    \item We propose a Progressive Localization Network (PLN) which progressively localizes the target moment in a coarse-to-fine manner. Through the progressive localization via multiple stages with diverse temporal granularities, PLN shows superiority in localizing short moments in long videos. To the best of our knowledge, this paper is the first work to localize the target moment progressively.

    \item We propose a conditional feature manipulation that adaptively manipulates the video clip features by referring to the learned knowledge of the previous stage, making the clip features more suitable for the more fine-grained localization in the later stage. Besides, we also introduce an upsampling connection to inject the learned coarse relevance of the previous stage into the later stage. All these components are beneficial to multi-stage localization.
    
    \item We conduct extensive experiments on three public datasets: TACoS \cite{regneri2013grounding}, ActivityNet Captions \cite{krishna2017dense} and Charades-STA \cite{gao2017tall}. Our PLN achieves a new state-of-the-art performance on TACoS, and compares favorably to state-of-the-art methods on ActivityNet Captions and Charades-STA. Besides, PLN is better at localizing short moments in long videos than recent one-stage models. 
\end{itemize}

\section{Related Work}
\subsection{Temporal Action Localization}
Given an untrimmed video, the task of temporal action localization is asked to localize segments where a pre-defined list of actions happen and predict the action label of localized segments. Owning to the success of deep learning in video understanding area, significant progress has been made in this task recently \cite{shou2016temporal,zeng2019graph,chen2019relation,chao2018rethinking,shou2017cdc,liu2020progressive,sst_buch_cvpr17,lin2019joint,yang2021deconfounded,yang2022video,liu2020reasoning}.
Existing efforts on temporal action localization could be roughly categorized into two groups: multi-stage approaches \cite{zeng2019graph,chao2018rethinking} and one-stage approaches \cite{lin2017single,sst_buch_cvpr17}.
For multi-stage approaches, they split the localization task into multiple steps mainly involving proposal generation, classifying whether actions of interest happen in proposals, and proposal boundary refinement. For instance, Shou \etal \cite{shou2016temporal} propose a multi-stage CNN model, which generates candidate segments, recognizes actions, and localizes temporal boundaries in three stages, respectively.
By contrast, one-stage approaches integrate multiple steps together, showing much more efficient inference. 
For instance, Liu \etal \cite{liu2021centerness} directly predict starting, ending and center of action proposal.
It is worth noting that our multi-stage PLN is essentially different from the above multi-stage models for temporal action localization.
In \cite{zeng2019graph,chao2018rethinking}, each stage plays a specific functional role and all stages have to be cascaded to generate the final localization results. By contrast, our model is able to perform localization in each stage, and the different stages focus on the distinct temporal granularities of input videos. 

\subsection{Language-based Moment Localization}

Language-based moment localization task is more challenging than temporal action localization.
The task not only requires understanding the video content~\cite{yang2016semantic,mei2013near} but also performs cross-modal reasoning~\cite{xu2021cross,peng2019cm} to predict the target moment.
As the pioneers for this task, Hendricks \etal \cite{hendricks17iccv} and Gao \etal \cite{gao2017tall} first generate candidate moments by sliding windows, and then determine the target moment according to the given natural language query.
In particular,  Hendricks \etal \cite{hendricks17iccv} propose to embed candidate moments with its context and queries into a shared space, where the candidate moment showing the highest relevance score with the given query is selected as the target moment.
Gao \etal \cite{gao2017tall} first fuse the candidate moment and query, and further employ a temporal regression network to produce their alignment scores and location offsets.

After that, this task has received increasing attention \cite{liu2018crossmodal,liu2018attentive,zhang2019mm,zhang2019cross,chenrethinking,hu2021video}.
Following the research line of \cite{gao2017tall}, a number of works \cite{Ge_2019_WACV,chen2019semantic} try to exploit the semantic concepts to improve the performance. Among them, instead of utilizing sliding windows to generate candidate moments, Chen \etal \cite{chen2019semantic} employ extracted semantic concepts from the input video to generate candidate moments with high probabilities. 
Similarly,  Xu \etal \cite{xu2019multilevel} first weight the video frames by their similarity to the given query, and select the frames that are more relevant to the query to build candidate moments.
Considering the above methods that ignore the temporal dependency of candidate moments, Zhang \etal \cite{zhang2019learning} construct the whole candidate moments into a 2D temporal feature map and then use a convolutional network to model the temporal dependencies of adjacent moments. As such temporal dependency of candidate moments is helpful, we also inherit it into our proposed multi-stage model.
As a follow-up work of \cite{zhang2019learning}, Liu \etal \cite{liu2021context} consider more candidate moments, and model all pairs of start and end indices within the video simultaneously with a biaffine mechanism. 
In~\cite{hu2021video}, Hu \etal introduce hashing techniques into the language-based moment localization task, which improves its efficiency and scalability in practice.

We also observe that some works utilize temporal anchors to localize the target moment without generating candidate moment in advance \cite{chen2018temporally,wang2019temporally,zhang2019man,zhang2019cross,lin2020moment,yuan2019semantic,yuan2020semantic,ning2021interaction}. 
These methods inherit the anchor ideas of object detection \cite{ren2015faster,redmon2018yolov3}, utilizing multiple pre-defined temporal anchors of different lengths. 
For instance, Chen \etal \cite{chen2018temporally} sequentially exploit the fine-grained frame-by-word interactions between video and sentence, and feed the fused features into binary classifiers to predict relevance scores of multiple pre-defined temporal  anchors at each time step.
In a follow-up work, Wang \etal \cite{wang2019temporally} additionally employ a boundary module to further explore the boundary-aware information, which leads to more precise localization.
Despite the good performance, the anchors need to be carefully designed for a specific dataset and the performance is sensitive to the sizes and number of anchors \cite{tian2019fcos}.

Recently, we notice a solution of formulating the localization task as an end-to-end regression problem \cite{lu2019debug, chenrethinking,Rodriguez_2020_WACV,zhang2020span,zeng2020dense,mun2020local}. For instance, Rodriguez \etal \cite{Rodriguez_2020_WACV}, Mun \etal \cite{mun2020local} and Liu \etal \cite{liu2021single} directly regress the starting and the ending of the target moment based on the fused video-query feature by an end-to-end model. 
In \cite{zeng2020dense}, Zeng \etal regress the distances from each frame to the starting and ending of the target moment described by the query.
With the similar idea of \cite{zeng2020dense}, Chen \etal \cite{chenrethinking} also regress the distance, while utilizes a frame feature pyramid to capture multi-level semantics.
Additionally, reinforcement learning which is well known to be effective for playing electronic games, is also applied to localize activity in videos \cite{wang2019language,Hahn2019tripping,wu2020tree,sun2021maban}. These works formulate the localization task as a sequence decision problem. As the most recent work using reinforcement learning, Sun \etal \cite{sun2021maban} propose to use two agents respectively adjust the start location and end location to move towards the target moment.
Additionally, we also observe that Cao \etal \cite{cao2020adversarial} apply adversarial learning paradigm in this task, which jointly optimize the performance of both video moment ranking and video moment localization.

As the task involves both videos and text, there are works that focus on devising more effective cross-modal interaction modules \cite{liu2018attentive,liu2018crossmodal,jiang2019cross,yuan2019to,zeng2020dense,chenhierarchical, xu2019multilevel}.
For example, Liu \etal \cite{liu2018crossmodal} propose a word attention based on the temporal context information in videos. Yuan \etal \cite{yuan2019to} design an elegant co-attention module to further exploit both video and query content. 
Zeng \etal \cite{zeng2020dense} devise a multi-level interaction module to fuse video and text using hierarchical feature maps.
Similarly, in \cite{lin2020moment}, Lin \etal propose a multi-stage cross-modal interaction, aiming to explore the intrinsic relations between video and query.
Our model is orthogonal to the improvement in cross-modal interaction, allowing us to flexibly embrace state-of-the-art cross-modal interaction modules.

Recently, we notice an increasing effort on solving the localizing task in a weakly-supervised manner\cite{mithun2019weakly,lin2020weakly,tan2021logan,wu2020reinforcement,zhang2020metal} or an unsupervised manner\cite{gao2021learning}. For instance, instead of utilizing annotated video-sentence pairs for model training, Gao \etal \cite{gao2021learning} resort to the existing visual concept detectors and a pre-trained image-sentence embedding model, and transfer the knowledge from the image domain to the video domain.
Though the weakly-supervised or unsupervised method are promising, their performance are still much worse than supervised methods. In this work, we mainly focus on the supervised methods.
Additionally, different from the above one-stage works, our paper is the first work for progressively localizing the target moment by multiple stages in a coarse-to-fine manner.
It is worth noting that our multi-stage PLN is different from previously proposed two-stage model \cite{xiao2021boundary}. In~\cite{xiao2021boundary}, Xiao \etal divide localization process into two stages, where they generate candidate moments in the first stage and then match the generated moments with the given sentence query in the second stage.
By contrast, in our work, we regard the one whole localization process as one stage. Besides, multiple stages progressively perform localization with distinct temporal granularities,  and are fused to predict the final result.

\section{Progressive Localization Network}
\begin{figure*}[tb!]
\centering\includegraphics[width=\linewidth]{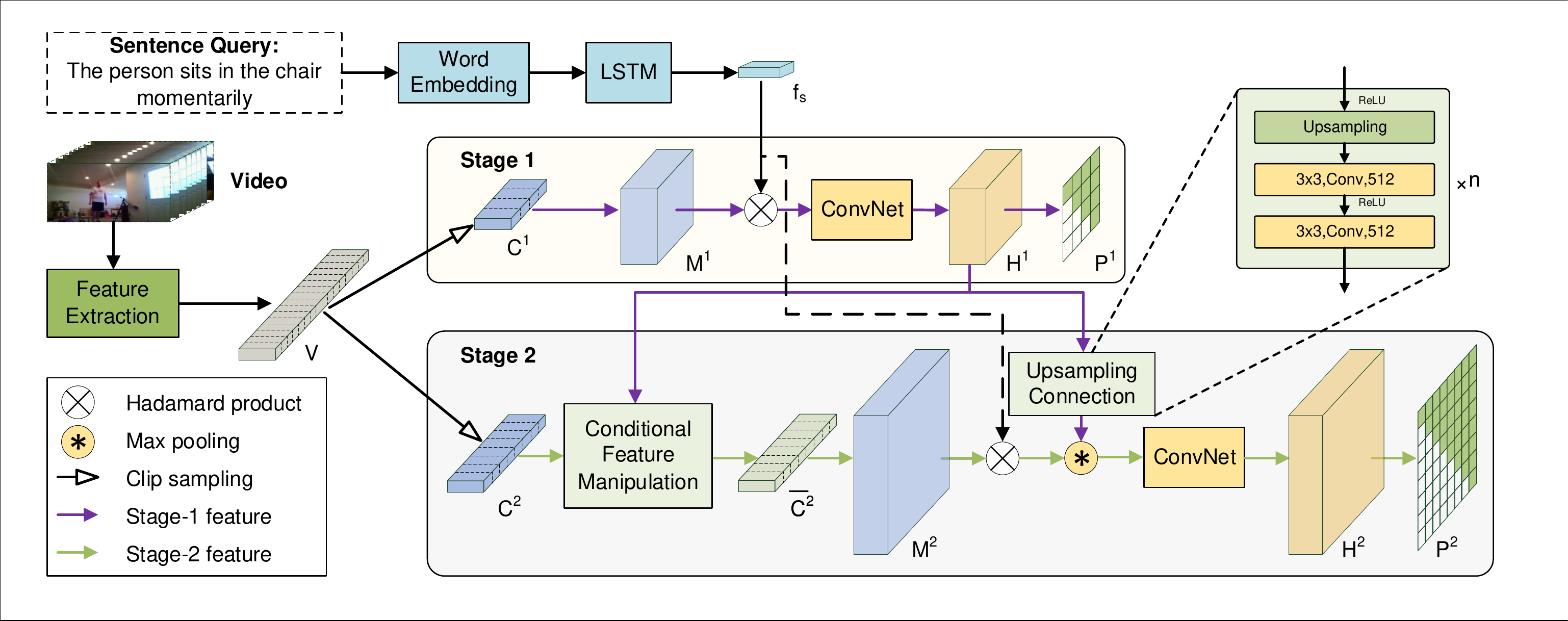}
\caption{The structure of our proposed two-stage Progressive Localization Network (PLN) which progressively localizes the target moment in a coarse-to-fine manner. Each stage has a localization branch, but with a distinct temporal granularity. 
The multiple localization branches are connected via Conditional Feature Manipulation (CFM) and Upsampling Connection (UC).
The CFM adaptively recalibrates the strengths of clip features under the guidance of the previous stage, and the UC allows the learned information of the previous stage into the later stage for more fine-grained localization.
}\label{fig:structure}
  \vspace{-0.15in}
\end{figure*}

Given an untrimmed video $v$ and a natural language sentence query $s$, the language-based moment localization task is required to localize a segment/moment which is semantically relevant to the given sentence query. As the task requires cross-modal reasoning and videos often contain intricate activities, it is not easy to directly localize the target moment in videos, especially for short moments in long videos. 
Therefore, we propose to progressively localize the target moment in a coarse-to-fine manner, and devise a model named as Progressive Localization Network (PLN).
PLN has multiple stages and each stage has a localization branch (Fig. \ref{fig:structure} illustrates the structure of two-stage PLN). 
In each stage $t(= 1, ..., T)$, a localization branch $g^t$ predicts a new localization result $\bf{P^t}$ according to the given video $v$, the sentence query $s$ and the learned information $\mathbf{H}^{t-1}$ of the previous stage:
\begin{equation}
\mathbf{P^t} = g^t(v,s,\mathbf{H}^{t-1}),
\end{equation}
where $\mathbf{H}^{t-1}$ indicates a feature map from the localization branch of the stage $t-1$, which contains the learned coarse relevance of candidate moments with the given query.
Such information benefits the later stage for more fine-grained localization and also connects the multiple localization branches in the whole model.
It is worth noting that the previous information in the first stage is unavailable, so we ignore the $\bf{H^{t-1}}$ for $g^1$ and only use the video $v$ and sentence $s$ as the inputs.
By combining the localization branches $g^1$ to $g^T$ together, our proposed model localizes the target moment progressively and more accurately.
In what follows, we first depict the input representation, following by the description of localization branch, conditional feature manipulation, training and inference details.

\subsection{Input Representation}

\textbf{Video Representation.}
Following the common practice~\cite{gao2017tall,chenrethinking,tan2021selective,yang2020tree}, we use the pre-trained CNN models to represent videos.
Concretely, given a video, we first split it into a fixed number of basic video units, where each unit consists of multiple consecutive frames.
For each unit, we extract the deep features using a CNN model pre-trained on ImageNet per frame, and mean pooling is performed to aggregate the features. 
In order to make the feature more compact, we further employ a fully connected (FC) layer with a ReLU activation.
Consequently, the video is described by a sequence of unit feature vectors $\mathbf{V}=\{\mathbf{u}_i\}_{i=1}^{l^v}$, where $\mathbf{u_i} \in \mathbb{R}^{d^{v}}$ is the feature vector of the $i$-th unit, $d^{v}$ denotes the dimensionality of the unit feature vectors and $l^v$ is the number of the video units.
It is worth noting that 3D CNNs, such as C3D~\cite{tran2015learning}, can also be used for feature extraction when treating consecutive frames as individual items.

\textbf{Sentence Query Representation.}
To represent sentence queries, we employ recurrent neural network which is known to be effective for modeling long-term word dependency in natural language text.
Specifically, given a sentence query $s$ consisting of ${l^s}$ words, we first embed each word of the given sentence into a word vector space by a pre-trained GloVe  model~\cite{pennington2014glove}, which results in a sequence of word embedding vectors $\{w_1, w_2, ... w_{l^s}\}$.
The word embedding vectors are then sequentially fed into a three-layer LSTM network, and the last hidden state vector of the last LSTM is taken as the representation of the sentence query, \ie ${\bf{f}}_s\in \mathbb{R}^{d^{s}}$, where $d^{s}$ is the hidden state size of the LSTM.

\subsection{Localization Branch}
In each localization branch, there is a component localizer which aims to localize relevant video moment under the guidance of the given natural language sentence query and learned coarse information of the previous stages. Here, we choose 2D-TAN \cite{zhang2019learning} as our component localizer, considering its state-of-the-art performance for language-based moment localization task. It is worth noting that theoretically other localization methods can also be used in our proposed progressive localization network.
As 2D-TAN is not specifically designed for the multi-stage localization manner, it can not be directly employed.
Hence, we adapt it to our proposed progressive multi-stage localization manner by additionally introducing a conditional feature manipulation and an upsampling connection layer. In what follows, we detail the localization branch in terms of candidate moment construction, relevance score prediction, and conditional feature manipulation.

\begin{figure}[tb!]
\centering\includegraphics[width=0.45\linewidth]{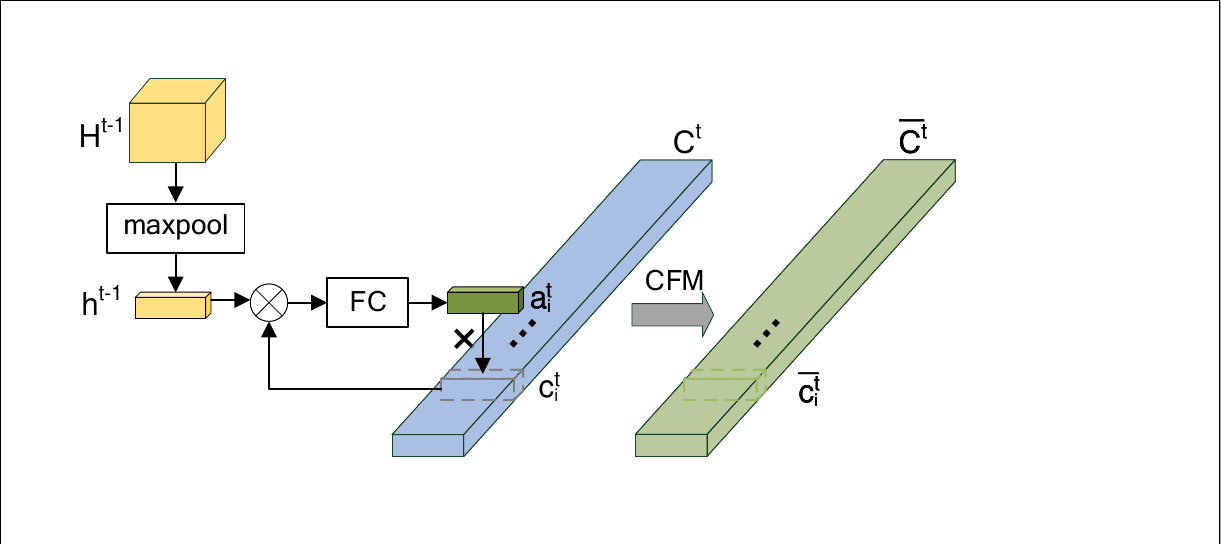}
\caption{The structure of conditional feature manipulation, which adaptively modulates the clip features under the guidance of the previous stage. 
}\label{fig:cfr}
\end{figure}

\textbf{Candidate Moment Construction.}
Given a video, this module generates candidate moments from the input video and then represent them into a 2D temporal feature map for further reasoning.
Concretely, in the $t$-th stage, given a video represented with a sequence of video unit feature vectors $\mathbf{V}=\{\mathbf{u}_i\}_{i=1}^{l^V}$, we split the video with a specific interval thus obtaining $N^t$ video clips $\{\mathbf{c}_i^t\}_{i=n}^{N^t}$, denoted as $\mathbf{C}^t$, where $\mathbf{c}_i^t$ indicates the representation of $i$-th generated video clip obtained by mean pooling over the corresponding feature vectors.
We perform the clip generation with different intervals in different stages resulting in video clips of the distinct temporal granularities, and make it satisfying $N^{t-1} < N^t$. The larger $N^t$ means the smaller temporal granularity of generated clips. 
In this way, the model structure conforms to the coarse-to-fine manner and is also helpful for locating relevant moment progressively. 
Additionally, in the stage $t(t>1)$, we also devise a Conditional Feature Manipulation (CFM) module to modulate the video clip features under the guidance of the previous stage, expecting to refine the clip features and make them more suitable for the more fine-grained localization in the later stage. 
The structure of CFM is illustrated in Fig. \ref{fig:cfr} and its detail will be introduced in the following section.

\begin{figure}[tb!]
\centering\includegraphics[width=0.98\linewidth]{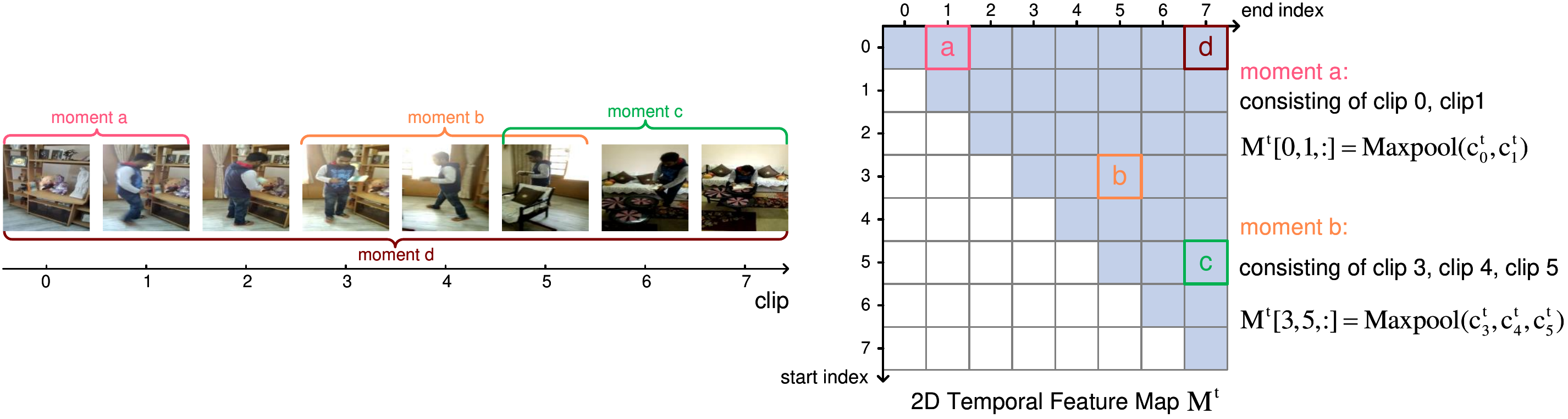}
\caption{An example of the 2D temporal feature map $\mathbf{M}^t$ built on a video of eight clips.
In $\mathbf{M}^t$, the first two axes indicate the start and end clip indexes respectively, and the third axis represents the moment features at the stage $t$. For simplicity, we omit the third axis in the figure.
The white ones are invalid, as valid moment should satisfy that the start index is smaller than the end index.
For the \textit{moment a} consisting of clip 0 and clip 1, its moment feature is $\mathbf{M}^t[0,1,:]$ obtained by employing max pooling over the clip features $\mathbf{c}_0^t$ and $\mathbf{c}_1^t$.
}\label{fig:fm}
\end{figure}

Based on the above generated video clips, we enumerate the contiguous clips to build candidate moments.
Theoretically, given a sequence of $N^t$ generated video clips, we can totally generate $\sum_{k=1}^{N^t}k$ different candidate moments with diverse lengths.
In order to facilitate the exploration of the temporal dependency between candidate moments, following the previous work~\cite{zhang2019learning}, we restructure the whole generated candidate moments to a 2D temporal feature map $\mathbf{M}^t\in \mathbb{R}^{N^t\times N^t\times d^V}$, where the first two axes indicate the start and end clip indexes at the stage $t$ respectively, and the third axis represents the moment features.
For example, $\mathbf{M}^t[i,j,:] = {\bf{m}}_{i,j}^t$ indicates the candidate moment that starting with the $i$-th video clip and ending with the $j$-th video clip, which is obtained by employing max pooling over the corresponding clip features, as exemplified in Fig.\ref{fig:fm}.
However, using all candidate moments will bring much more redundant information. To alleviate it, we use a sparse sampling strategy~\cite{zhang2019learning} that removes the redundant candidates containing large overlaps with the selected candidates. Briefly, we densely sample moments of short duration, and gradually increase the sampling interval when the moment duration becomes long. The restructured 2D temporal feature map $\mathbf{M}^t$ is further used for the relevance score prediction.

\textbf{Relevance Score Prediction.}
Given a list of candidate moments represented as a feature map $\mathbf{M}^t$ and a query represented as ${\bf{f}}_s$, this module aims to predict the relevance score map $ \mathbf{P^t} \in \mathbb{R}^{N^t\times N^t}$, where $\mathbf{P^t}[i,j]$ indicates the relevance of the candidate moment starting with the $i$-th video clip and ending with the $j$-th video clip with the given query.

Concretely, we first project $\mathbf{M}^t$ and $\mathbf{f}_s$ into a $d^u$-dim unified space by a Fully Connected (FC) layer respectively, and fuse them through an element-wise multiplication, resulting in the fused feature map $\mathbf{F}^t$.
Besides, we also consider the learned information in the previous stage crucial for the final prediction, thus employ a connection to inject the coarse relevance information of the previous stage into the follow-up stage. 
Notice that there is no previous stage in the first stage. 
Formally, after the injection, we obtain the fused feature that absorbed the information of the previous stage:
\begin{equation}
    \begin{aligned}
        \mathbf{G}^t&=
        \begin{cases}
        \mathbf{F}^t & \text{if t $=$ 1,}\\
        \mathbf{F}^t \circledast upsample(\mathbf{H}^{t-1}) & \text{if t $>$ 1,}\\ 
        \end{cases}\\
    \end{aligned}
\end{equation}
where $\circledast$ denotes the element-wise max pooling, $upsample$ indicates our devised upsampling connection, and $\mathbf{H}^{t-1}$ is a feature map of the last convolutional layer in the previous stage. Here we can not directly employ a skip connection, as the sizes of $\mathbf{F}^t$ and $\mathbf{H}^{t-1}$ are different due to the different temporal granularities in different stages. Therefore, we devise a new upsampling connection which is implemented by $n$ stacked upsampling blocks, and each block has an upsampling layer with a factor of 2 and two stacked convolutional layers with $3\times3$ kernels, as shown in the upper right of Fig. \ref{fig:structure}.
Then a convolutional network is further utilized to exploit the candidate dependency based on the fused feature map $\mathbf{G}^t$.
The convolutional network is comprised of 2 convolutional layers with $5\times5$ kernels, and each layer is followed by a ReLU activation.
The output of the convolutional network in the $t$-th stage is denoted as $\bf{H}^t$.
Finally, we feed the output $\bf{H}^t$ into a FC layer
to predict the final relevance score map $ \mathbf{P}^t \in \mathbb{R}^{N\times N}$:
\begin{equation}
    \mathbf{P}^t = sigmoid (\mathbf{W}^t \cdot \mathbf{H}^t+\mathbf{b}^t),
\end{equation}
where $sigmoid$ indicates an element-wise sigmoid activation, $\bf{W}^t$ denotes the affine matrix and $\bf{b}^t$ is the bias term. In our implementation, the FC layer is implemented with a 1x1 convolutional layer.

\textbf{Conditional Feature Manipulation.} \label{subsec:cfm}
As shown in Fig. \ref{fig:cfr}, given a sequence of video clip feature vectors $\mathbf{C}^t = \{\mathbf{c}_i^t\}_{i=n}^{N^t}$, CFM modulates the given features under the guidance of the previous stage, obtaining the manipulated features as $\overline{\mathbf{C}^t}=\{\overline{\mathbf{c_i}^t}\}_{i=1}^{N^t}$. 

Specifically, for each clip feature vector $\mathbf{c}_i^t$ in $\mathbf{C}^t$, we modulate it to $\overline{\mathbf{c}_i^t}$ with respect to the feature map $\mathbf{H}^{t-1}$.
First of all, a max pooling layer is employed to aggregate $\mathbf{H}^{t-1}$ to a feature vector $\mathbf{h}^{t-1}$.
Afterwards, we fuse $\mathbf{h}^{t-1}$ and $\mathbf{c}_i^t$ through element-wise multiplication and then a FC layer with the $sigmoid$ activation is employed to generate a scaling vector $\mathbf{a}_i^t \in \mathbb{R}^{d^{S}}$:
\begin{equation}
\mathbf{a}_i^t = sigmoid(\mathbf{W}_r^t \cdot (\mathbf{h}^{t-1} \odot \mathbf{c}_i) + \mathbf{b}_r^t),
\end{equation}
where $\mathbf{W}_r^t$ and $\mathbf{b}_r^t$ parameterize the FC layer, $\odot$ indicates the Hadamard product. 
With the above generated scaling vector $\mathbf{a}^t$, the clip feature vector $\mathbf{v}_i$ is manipulated by scaling as:
\begin{equation}
 \overline{\mathbf{c}_i}^t  = \mathbf{c}_i \odot \mathbf{a}_i^t.
\end{equation}
Our CFM is inspired by FiLM \cite{perez2018film} which manipulates a neural network’s intermediate features for visual reasoning.
In our PLN, the motivation for CFM is two-fold: 
(i) we wish to adaptively recalibrate the clip features, thus making the clip features more suitable for the more fine-grained localization in the later stage. 
(ii) we wish to make clip features diverse across the stages thus promoting the localization branches more complementary.

After performing the conditional feature manipulation,  the clips of the given video can be represented as $\overline{\mathbf{C}^t}=\{\overline{\mathbf{c_i}^t}\}_{i=1}^{N^t}$, which is used in the later stages $t(t>1)$ to replace the original video clip features $\mathbf{C}^t$. 

\subsection{Training and Inference}

\textbf{Training.}
To train the proposed PLN, we add a classification loss per localization branch, and multiple losses of all branches are jointly considered.
Specifically, in the $t$-th stage, a cross-entropy loss is employed, which is defined as:
\begin{equation}\label{eq:loss}
\mathcal{L}_t = - \frac{1}{V^t}\sum_{i=1}^{V^t} (y_i^t \log p_i^t + (1-y_i^t) \log(1-p_i^t)),
\end{equation}
where $p_i^t \in \mathbf{P}^t$ is the predicted relevance score of the $i$-th candidate moment and $V^t$ denotes the total number of valid candidates in the $t$-th stage.
For supervision labels, we employ soft labels based on candicate moments' temporal Intersection over Union (IoU) with the ground-truth moment instead of hard binary labels.
Specifically, the label $y_i^t$ in the $t$-th stage is defined as:
\begin{equation}
\begin{aligned}\label{eq:iou}
    y_i^t& =
    \begin{cases}
    0 & {{o_i^t} \le \tau}, \\
    \frac{o_i^t-\tau}{{1-\tau}} & { {o_i^t} >  \tau}, 
    \end{cases}
\end{aligned}
\end{equation}
where $o_i^t$ denotes IoU of the corresponding candidate moment with the ground-truth moment, $\tau$ indicates the IoU threshold. Recall that candidate moments are generated with clips of different intervals in different stages, so $V^t$ and $y_i^t$ are different across the stages.

Finally, the joint loss of our proposed model with $T$ stages is defined as:
\begin{equation}\label{eq:final_loss}
\mathcal{L} =\sum_{t=1}^{T} \lambda_t \mathcal{L}_t,
\end{equation}
where $\lambda_t$ indicates trade-off coefficient for the $t$-th stage.
The PLN is trained by minimizing the Eq.\ref{eq:final_loss} in an end-to-end manner except the CNN used for extracting video features and the GloVe for word embedding are pre-trained and fixed. 

\textbf{Inference.}
After the model being trained, each localization branch is able to predict a relevance score map. 
Depending on which relevance score map is used for final localization, here we consider two strategies:

\textit{Strategy 1}. 
In this strategy, we select one score map among all the predicted score maps, while ignoring the others.
According to the selected score map, all the corresponding candidate moments will be ranked and Non Maximum Suppression (NMS) strategy is further used to remove the redundant predicted moments. 

\textit{Strategy 2}.
Considering all the predicted relevance score maps might be valuable for localization, we ensemble all the predicted score maps to obtain the final fused score map.
Specifically, if a candidate moment is associated with multiple score maps, we average their corresponding scores in all stages as the final fused relevance score.
Similarly, all the corresponding candidate moments are ranked in terms of their fused score and the NMS strategy is also performed.

\section{Evaluation}

In this section, we first introduce our experimental settings, then compare our proposed PLN with its best setup against the state-of-the-art on three datasets.
Furthermore, we conduct extensive ablation studies to investigate the impact of major design choices in the proposed model. Finally, we analyse the results on different types of moments.

\begin{table} [tb!]
\renewcommand{\arraystretch}{1.2}
\caption{Brief statistics of three datasets used in our experiments.}
\label{tab:dataset_statistics}
\centering 
\scalebox{0.8}{
\begin{tabular}{l*{8}{c}}
\toprule
\multirow{2}{*}{\textbf{Datasets}}   &
\multirow{2}{*}{\textbf{Domain}}   &
\multicolumn{3}{c}{\textbf{\#Videos}} && \multicolumn{3}{c}{\textbf{\#Sentences}}  \\
 \cmidrule{3-5} \cmidrule{7-9}
&& train & val & test  && train & val & test \\
\toprule
\textbf{TACoS}                    &Cooking  & 75     & 27    & 25     && 9790     & 4436  & 4001\\
\textbf{ActivityNet Captions}     &Open  & 10009  & 4917  & 4885   && 37421    & 17505 & 17031 \\
\textbf{Charades-STA}             &Indoors  & 5336   & -     & 1334   && 12408    & -     & 3720  \\
\bottomrule
\end{tabular}
 }
\end{table}

\begin{figure}[tb!]
\centering
\subfigure[]{
\includegraphics[width=0.3\linewidth]{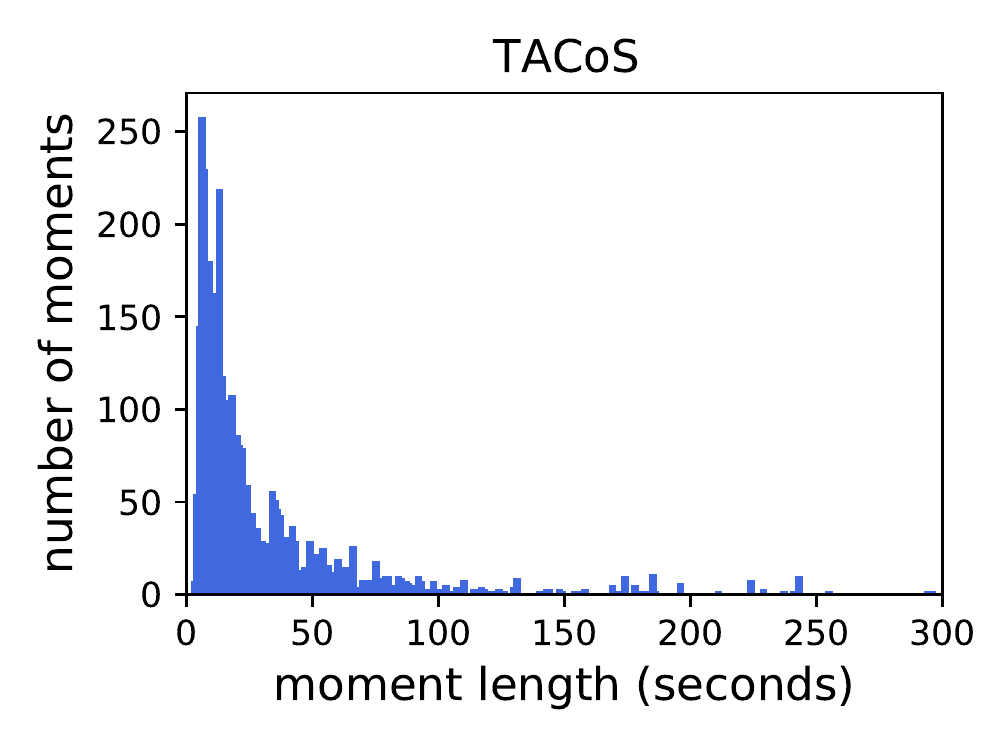}
}
\subfigure[]{
\includegraphics[width=0.3\linewidth]{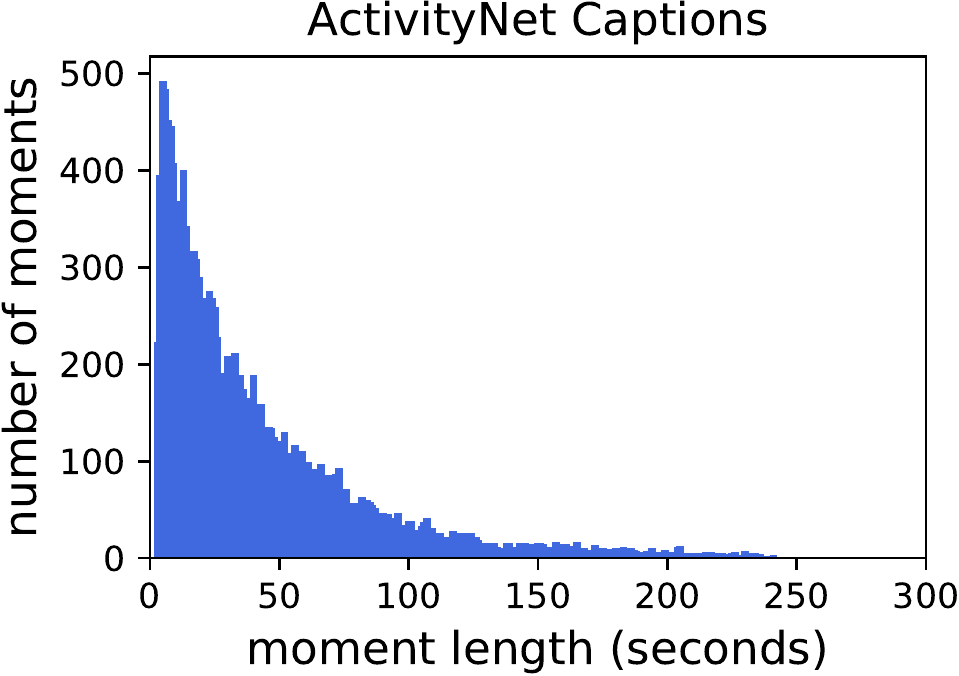}
}
\subfigure[]{
\includegraphics[width=0.3\linewidth]{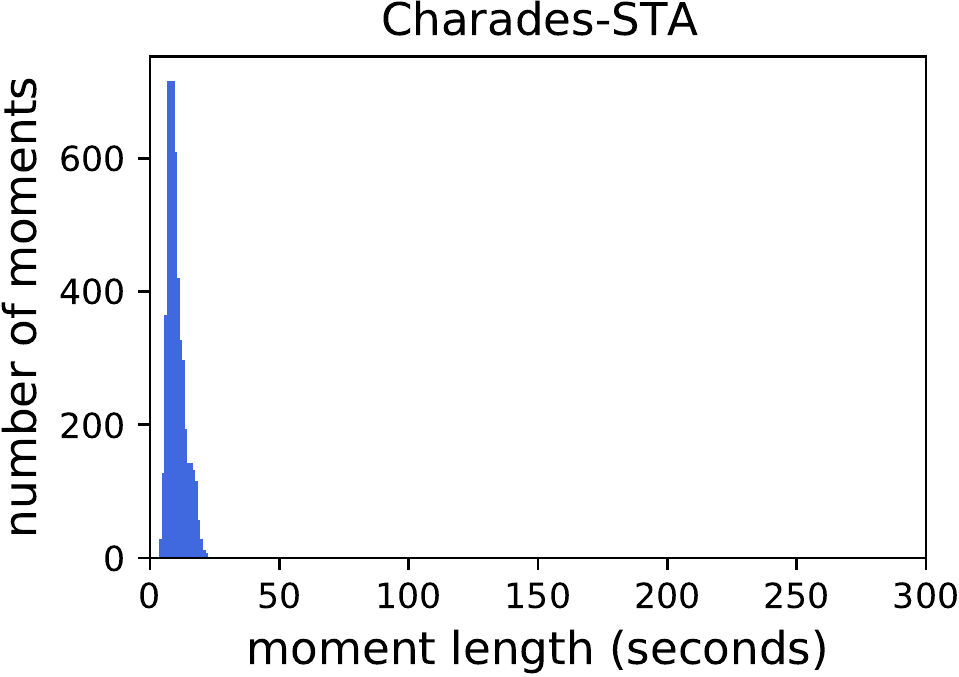}
}
\vspace{-3mm}
\caption{The distributions of target moment lengths on TACoS, ActivityNet Captions and Charades-STA. TACoS and ActivityNet Captions show much larger variation in the moment lengths than Charades-STA.}
\label{fig:moment_group_num}
\end{figure}

\subsection{Experimental Settings}

\subsubsection{Datasets}

We conduct extensive experiments on three commonly used language-based moment localization datasets: TACoS,  ActivityNet Captions and Charades-STA. Table \ref{tab:dataset_statistics} summarizes the brief statistics of these datasets, and Fig. \ref{fig:moment_group_num} shows the distributions of target moment lengths on the three datasets.

\textit{TACoS} \cite{regneri2013grounding} is built based on the MPII Cooking Composite Activities video dataset \cite{rohrbach2012script}, consisting of 127 videos about activities that happened in the kitchen. 
The average length of videos is about 286 seconds, and the average length of target moments is about 27 seconds. Among them, lots of queries are about short fine-grained actions, such as \textit{takes out the knife}. Such short spanned moments show the challenging nature of this dataset.
We follow the setup in 2D-TAN\cite{zhang2019learning}, with $9790$, $4436$ and $4001$ sentence-moment pairs for training, validation and testing, respectively.

\textit{ActivityNet Captions} \cite{krishna2017dense} is the largest dataset for the task of moment localization with natural language, which is built on ActivityNet v1.3 dataset \cite{caba2015activitynet}.
The dataset is originally developed for video captioning, and now popular for the task of moment localization with natural language as these two tasks are reversible.
Compared with the above two datasets, the videos are open domain and more diverse in content.
It totally has $19811$ videos, and the average length is about 117 seconds.
For the fair comparison, we follow the dataset split used in \cite{zhang2019cross,zhang2019learning} using $37417$, $17505$, and $17031$ sentence-moment pairs for training, validation, and testing, respectively.

\textit{Charades-STA} \cite{gao2017tall} is built on the top of the Charades dataset~\cite{sigurdsson2016hollywood} which is introduced for action recognition and localization. The Charades dataset has $9,848$ videos about daily indoor activities and each video is annotated with the video-level descriptions. The videos are 30 seconds long on average.
Gao \etal \cite{gao2017tall} extend this by including the sentence-level temporal annotation to create Charades-STA which has $12408$ sentence-moment pairs for training and $3720$ pairs for testing.

\subsubsection{Evaluation Metric}
Following the previous work~\cite{chen2018temporally}, we report the performance in terms of the metrics $Rank@n,IoU=m$ ($n\in\{1, 5\}$, $m\in\{0.1,0.3,0.5,0.7\}$) and $mIoU$. $Rank@n,IoU=m$ is defined as the percentage of test queries for which at least one relevant moment with higher IoU than $m$ is found among the top-$n$ retrieved results.
$mIoU$ is the mean IoU of top-1 prediction result with the ground truth moment over all test queries. All results are reported in percentage (\%).

\subsubsection{Implementation Details}
Here we detail common implementations of our proposed model in terms of model structure, training and inference. 
For the sentence encoding, the size of hidden states in the LSTM is 512, so $d^s=512$. 
For the video features, we utilize the same visual features provided by \cite{zhang2019learning}, \ie 4096-dim C3D feature~\cite{tran2015learning} on TACoS, 500-dim C3D feature on ActivityNet Captions, and 4096-dim VGG feature~\cite{Simonyan15} on Charades-STA. 
Besides, we notice a trend of using a stronger I3D feature on Charades-STA, so we also use the 1024-dim I3D feature provided by \cite{Rodriguez_2020_WACV}.
The parameters of the convolution networks (ConvNet) in all stages are shared.
For the upsampling connection, we use 2 upsampling blocks for two-stage models(\ie $n$=$2$), and 1 upsampling blocks for three-stage and four-stage models(\ie $n$=$1$).
We use two-stage PLN to compare the state-of-the-art models, and the number of sampled clips $N^t$ is set to 32 and 128 on TACoS, 16 and 64 on ActivityNet Captions and Charades-STA.
On ActivityNet Captions, we additionally add a sinusoidal positional encoding~\cite{vaswani2017attention} to a sequence unit features, thus boost the order of the sequence.
For the model training, we empirically set the loss weights $\lambda_t$ to 1.0 and 1.5 for the two-stage model, 1.0, 1.3 and 1.5 for the three-stage model, and 1.0, 1.2, 1.5 and 2.0 for the four-stage model. 
The $IoU$ threshold $\tau$ with ground-truth moment is set to be 0.5.
Additionally, we use Adam optimizer~\cite{kingma2014adam} to train the model. The initial learning rate and the batch size are empirically set to be $0.0001$ and $32$ respectively. The maximum number of epochs is 50. 
For model inference, we use non maximum suppression (NMS) strategy to remove the redundant predicted moments. Following the previous work \cite{zhang2019learning}, the threshold of NMS is set to be 0.4 on TACoS, 0.5 on ActivityNet Captions and 0.45 on the Charades-STA dataset.

\begin{table*}[tb!]
\caption{Performance comparison on TACoS and ActivityNet Captions, where all the models use the same C3D feature. As the scores of LGI on TACos are unavailable on the original paper, we have re-trained it using its public code with the same video feature. Other scores are directly cited from the original papers. 
The best and second-best results are in \textbf{bold} and \underline{underlined}, respectively.
}
\label{tab:test11}
\centering
\scalebox{0.7}{
\begin{threeparttable}
\begin{tabular}{@{}l*{19}{c} @{}}
\toprule
\multirow{3}{*}{\textbf{Method}} &
  \multicolumn{8}{c}{\textbf{TACoS}} &&
  \multicolumn{8}{c}{\textbf{ActivityNet Captions}} &
   \\ \cmidrule{2-9} \cmidrule{11-18}
   &
  \multicolumn{3}{c}{\textbf{Rank@1,IoU=m}} &&
  \multicolumn{3}{c}{\textbf{Rank@5,IoU=m}} &
  \multirow{2}{*}{\textbf{mIoU}} &&
  \multicolumn{3}{c}{\textbf{Rank@1,IoU=m}} &&
  \multicolumn{3}{c}{\textbf{Rank@5,IoU=m}} &
  \multirow{2}{*}{\textbf{mIoU}} & \\ \cmidrule{2-4} \cmidrule{6-8} \cmidrule{11-13} \cmidrule{15-17}
         &                  0.1   & 0.3   & 0.5   && 0.1   & 0.3   & 0.5   &&      & 0.3   & 0.5   & 0.7   && 0.3   & 0.5   & 0.7   &       &  \\ 
\midrule
CTRL\cite{hendricks17iccv}             & 24.32 & 18.32 & 13.30 && 48.73 & 36.69 & 25.42 & 11.98 && -     & -     & -     && -     & -     & -     & -     &  \\
ACRN\cite{liu2018attentive}       & 24.22 & 19.52 & 14.62 && 47.42 & 34.97 & 24.88 & -     && 49.70 & 31.67 & 11.25 && 76.50 & 60.34 & 38.57 & -     &  \\
TGN\cite{chen2018temporally}      & 41.87 & 21.77 & 18.90 && 53.40 & 39.06 & 31.02 & 17.93 && -     & -     & -     && -     & -     & -     & -     &  \\
ACL-K\cite{Ge_2019_WACV}         & 31.64 & 24.17 & 20.01 && 57.85 & 42.15 & 30.66 & -     && -     & -     & -     && -     & -     & -     & -     &  \\
CMIN\cite{zhang2019cross}         & 32.48 & 24.64 & 18.05 && 62.13 & 38.46 & 27.02 & -     &&\underline{63.61} & 43.40 & 23.88 && 80.54 & 67.95 & 50.73 & -     &  \\
DEBUG\cite{lu2019debug}           & 41.15 & 23.45 & -     && -     & -     & -     & 16.03 && 55.91 & 39.72 & -     && -     & -     & -     & 39.51 &  \\
TripNet\cite{Hahn2019tripping}        & -     & 23.95 & 19.17 && -     & -     & -     & -     && 48.42 & 32.19 & 13.93 && -     & -     & -     & -     &  \\
ABLR\cite{yuan2019to}                  & 34.70 & 19.50 & 9.40  && -     & -     & -     & 13.40 && 55.67 & 36.79 & -     && -     & -     & -     & 36.99 &  \\
QSPN-Cap\cite{xu2019multilevel}      & -     & -     & -     && -     & -     & -     & -     && 45.30 & 27.70 & 13.60 && 75.70 & 59.20 & 38.30 & -     &  \\
SLTA\cite{jiang2019cross}             & 23.13 & 17.07 & 11.92 && 46.52 & 32.90 & 20.86 & -     && -     & -     & -     && -     & -     & -     & -     &  \\
SCDM\cite{yuan2019semantic}           & -     & 26.11 & 21.17 && -     & 40.16 & 32.18 & -     && 54.80 & 36.75 & 19.86 && 77.29 & 64.99 & 41.53 & -     &  \\
VSLNet\cite{zhang2020span}        & -     & 29.61 & 24.27 && -     & -     & -     & 24.11 && 63.16 & 43.22 & 26.16 && -     & -     & -     & 43.19 &  \\
CBP\cite{wang2019temporally}           & -     & 27.31 & 24.79 && -     & 43.64 & 37.40 & 21.59 && 54.30 & 35.76 & 17.80 && 77.63 & 65.89 & 46.20 & 36.85 &  \\
GDP\cite{chenrethinking}                & 39.68 & 24.14 & -     && -     & -     & -     & 16.18 && 56.17 & 39.27 & -     && -     & -     & -     & 39.80  &  \\
TSP-PRL\cite{wu2020tree}                  & -     & -     & -     && -     & -     & -     & -     && 56.08 & 38.76 & -     && -     & -     & -     & 39.21 &  \\

2D-TAN\cite{zhang2019learning}        & 47.59 & 37.29 & 25.32   && \underline{74.46} & \underline{57.81} & 45.04  & \underline{25.19}  && 59.45 & 44.51 & \underline{27.38} && \underline{85.65} & \underline{77.13} & \underline{62.26} &43.29   \\

PMI-LOC\cite{chen2020learning}                & -     & -     & -     && -     & -     & -     & -     && 59.69 & 38.28 & 17.83 && -     & -     & -     & - &  \\
DRN\cite{zeng2020dense}                    & -     & -     & 23.17 && -     & -     & 33.36 & -     && -     & \underline{45.45} & 24.36 && -     & \textbf{77.97} & 50.30  & -     &  \\
LGI\cite{mun2020local}                  & \underline{50.69}  &\underline{37.52} &22.74   &&-     &-     & -     &24.37     && 58.52 & 41.51 & 23.07 && -     & -     & -     & 41.13 &  \\ 
SCDM\cite{yuan2020semantic}  &- &27.64 &23.27 &&- &40.06 &33.49 &- &&55.25 &36.90 &20.28 &&78.79 &66.84 &42.92  &- \\
ABIN\cite{zhang2020temporal}  &31.84 &25.28 &19.70 &&- &- &- &20.12 &&63.19 &44.02 &24.63 &&- &- &-  &\textbf{45.08} \\
CMIN\cite{lin2020moment}   &36.88 &27.33 &19.57 &&64.93 & 43.35 & 28.53 &- &&\textbf{64.41} &44.62 &24.48 && 82.39 &69.66 &52.96 & - & \\
BPNet\cite{xiao2021boundary}  & - &25.96 & 20.96 && - & -     & -     & 19.53  && 58.98 & 42.07 & 24.69 && -     & -     & -     & 42.11 &  \\ 
I2N\cite{ning2021interaction}   &- &31.47 &\underline{29.25} &&- & 52.65 & \underline{46.08} &- &&- & - &- && - &- &- & - & \\
MABAN\cite{sun2021maban}   &- &- &- &&- & - & - &- &&- &44.88 &25.66 && - &- &- & - & \\
CMHN\cite{hu2021video}   &- &30.04 &25.58 &&- & 44.05 &35.23 &- &&62.49 &43.47 &24.02 &&85.37 &73.42 &53.16  &- & \\


\textit{PLN}  & \textbf{53.74} & \textbf{43.89} & \textbf{31.12} && \textbf{75.56} &\textbf{65.11} & \textbf{52.89} & \textbf{29.70} && 59.65     & \textbf{45.66}     & \textbf{29.28}     && \textbf{85.66}   & 76.65    & \textbf{63.06}    &\underline{44.12}    \\
\bottomrule
\end{tabular}
\end{threeparttable}
}
\end{table*}

\subsection{Comparison to the State-of-the-Arts}

\subsubsection{Quantitative Comparison}
In this section, we compare our best model with the state-of-the-art methods. The analysis on our model with different setting are summarized in Section \ref{ssec:ablation}.
Table~\ref{tab:test11} shows the performance comparison on both TACoS and ActivityNet Captions, where all the models use the same C3D feature.
Besides, except for our proposed PLN, all other models are one-stage. 
Our proposed multi-stage PLN consistently achieves the best performance on TaCoS, and compares favorably to existing methods on ActivityNet Captions.
Moreover, our multi-stage PLN in a coarse-to-fine manner outperforms 2D-TAN~\cite{zhang2019learning} with a clear margin. 2D-TAN  can be deemed as the degraded version of our proposed model, which only has one localization branch with clips of a fixed temporal granularity as input. 
The results again justify the viability of localizing the target moment via multiple stages with diverse temporal granularities.
Additionally, we also notice a phenomenon that the performance improvement of our PLN over 2D-TAN is more significant on TACoS than that on ActivityNet Captions.
As the moment length proportion of the entire video on TACoS is much smaller than that on ActivityNet Captions, we conclude that our PLN with multiple stages is more beneficial for localizing relatively short moments in long videos.

\begin{table}[tb!]
  \caption{Performance comparison with the same VGG or I3D feature on Charades-STA. 
        The best and second-best results are in \textbf{bold} and \underline{underlined}, respectively.}
  \label{tab:charades-temp}
    \centering
  \scalebox{0.78}{
  \begin{threeparttable}
  
    \begin{tabular}{@{}l*{9}{c} @{}} 
      \toprule
      \multirow{2}{*}{\textbf{Method}} &
      \multicolumn{3}{c}{\textbf{Rank@1,IoU=m}} &&
      \multicolumn{3}{c}{\textbf{Rank@5,IoU=m}} &
      \multirow{2}{*}{\textbf{mIoU}} \\
      \cmidrule{2-4} \cmidrule{6-8}
      & 0.3   & 0.5   & 0.7     &&  0.3   & 0.5   & 0.7   &       \\ 
      \midrule
      
        \textbf{VGG:}  & &       &       &       &&     &       &       &       \\
        SAP\cite{chen2019semantic}          & -     & 27.42 & 13.36 && -   & 66.37 & 38.15 & -     \\
        SM-RL\cite{wang2019language}     & -     & 24.36 & 11.17 && -   & 61.25 & 32.08 & -     \\
        2D-TAN\cite{zhang2019learning}     & \underline{57.31}    & 42.80 & 23.25 && \underline{93.49}   & 83.84 & 54.17 & \underline{39.23}     \\
        DRN\cite{zeng2020dense}         & -     & \underline{42.90}  & \underline{23.68} && -   & \textbf{87.80}  & \underline{54.87} & -     \\

        \textit{PLN}    & \textbf{59.33} & \textbf{45.43} & \textbf{26.26}   &&  \textbf{95.48} & \underline{86.32} & \textbf{57.02} & \textbf{41.28} \\
        
        \midrule
        \textbf{I3D:} &  &     &       &       &&     &       &       &       \\
        MAN\cite{zhang2019man}      & -     & 46.53 & 22.72 && -   & 86.23 & 53.72 & -     \\
        SCDM\cite{yuan2019semantic}        & -     & 54.44 & 33.43 && -   & 74.43 & 58.08 & -     \\
        TMLGA\cite{Rodriguez_2020_WACV}  & 67.53 & 52.02 & 33.74 && -   & -     & -     & -     \\
        DRN\cite{zeng2020dense}       & -     & 53.09 & 31.75 && -   & \textbf{89.06} & \underline{60.05} & - \\ 
        SCDM\cite{yuan2020semantic}  &- &54.92 &34.26 &&- &76.50 &60.02 & - \\
        LGI   \cite{mun2020local}    & \textbf{72.96} & \textbf{59.46} & \textbf{35.48} && -   & -     & -     & \textbf{51.38}    \\
        BPNet   \cite{xiao2021boundary}      & 65.48 & 50.75 & 31.64 && -   & -     & -     & 46.34 \\
        I2N\cite{ning2021interaction}  & - &\underline{56.61} &34.14 &&- &81.48 & 55.19 & - \\ [2pt]
        \textit{PLN}     & \underline{68.60} & 56.02 & \underline{35.16}   &&  \textbf{94.54} & \underline{87.63} & \textbf{62.34} & \underline{49.09} \\ 
        \bottomrule
        \end{tabular}
        \end{threeparttable}
    }
\end{table}

The performance comparison on Charades-STA is shown in Table \ref{tab:charades-temp}.
With the same VGG feature, our PLN performs the best except in terms of Rank@5 with IoU of 0.5. The result again confirms the effectiveness of PLN.
Additionally, with the same I3D feature, PLN is worse than LGI~\cite{mun2020local} on Charades-STA. 
We conjecture it to that LGI explores local-global video-text interactions via temporal attention for video-text fusion, while PLN utilizes a relatively simple element-wise multiplication.
Moreover, LGI is a regression-based method, which is good at fitting data of small variations. Hence, LGI performs better on Charades-STA where the temporal length variation of target moments is small (see Fig.~\ref{fig:moment_group_num}).
However, on TACoS and ActivityNet Captions where the temporal length variations of target moments are much larger, our proposed PLN consistently performs better.
The results allow us to conclude that our proposed PLN is better to handle moments with large variations in temporal lengths. We attribute it to that PLN has multiple stages for localization and different stages are responsible for the candidate moments generated with the clips of distinct temporal granularities.

\begin{figure*}[b!]
\centering\includegraphics[width=\linewidth]{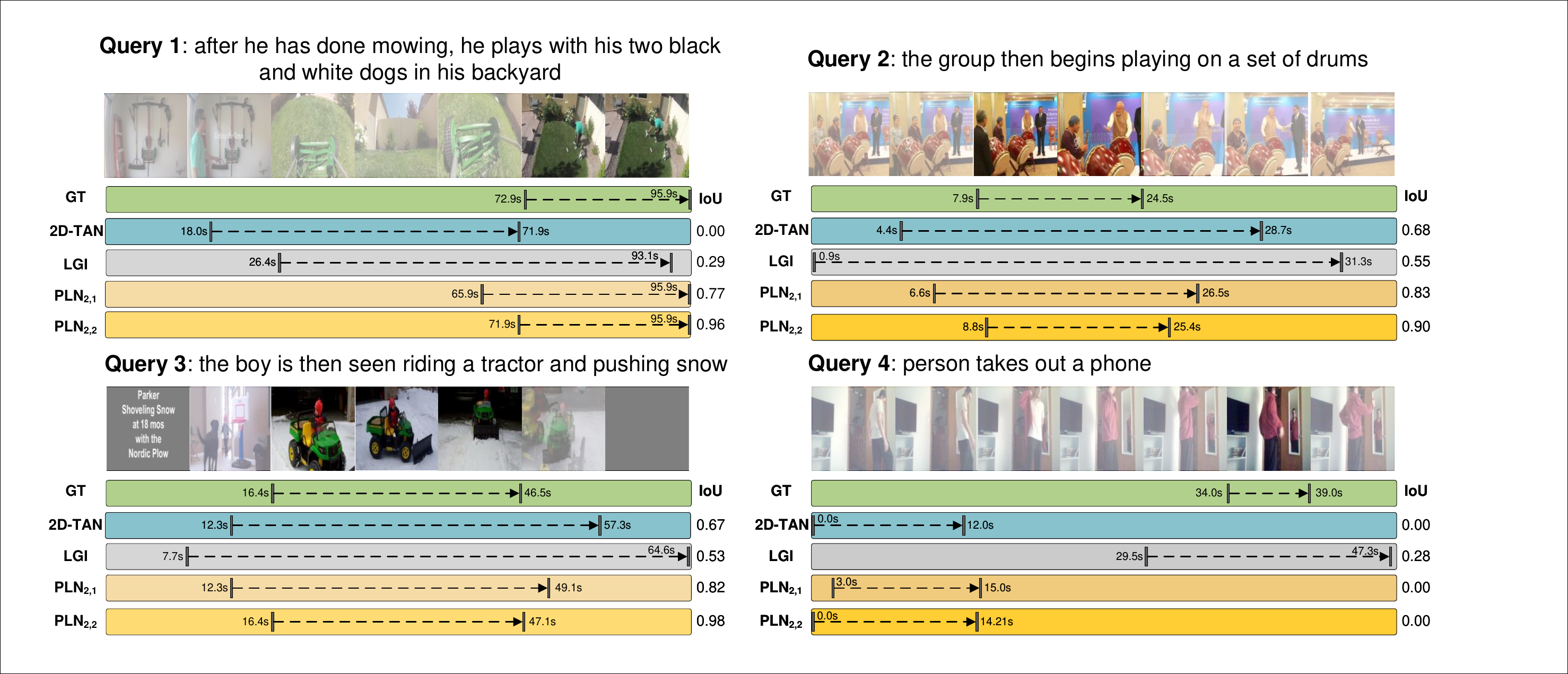}
 \vspace{-0.25in}
  \caption{
  Qualitative comparison of our proposed PLN with 2D-TAN~\cite{zhang2019learning} and LGI~\cite{mun2020local}. GT denotes the ground-truth result. 
  PLN$_{2,1}$ indicates our proposed two-stage PLN using the output of stage 1 for prediction, while PLN$_{2,2}$ is the same two-stage model but uses the output of stage 2. }
  
  \label{fig:good_example}
  \vspace{-0.15in}
\end{figure*}

\begin{figure*}[tb!]
\centering\includegraphics[width=\linewidth]{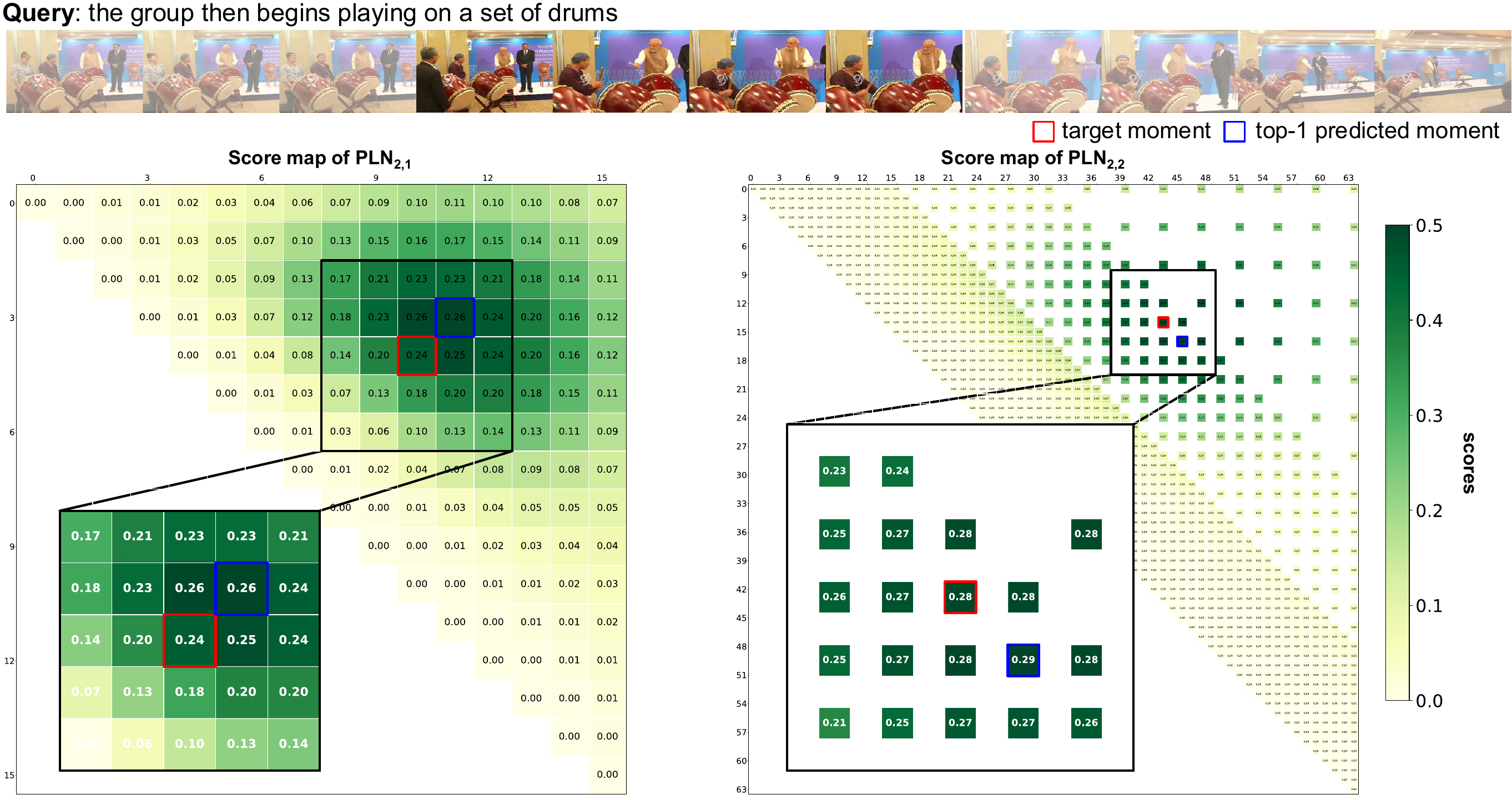}
 \vspace{-0.25in}
  \caption{An example of language-based moment localization by our model on ActivityNet Captions. 
  For each query and the corresponding video, we illustrate the predicted score maps of PLN$_{2,1}$ and PLN$_{2,2}$, where red boxes indicate target moments and blue boxes indicate top-1 predicted moments.
  Best viewed in zoom in. More examples are demonstrated in the supplementary materials.}
  \label{fig:vis}
  \vspace{-0.15in}
\end{figure*}

\subsubsection{Qualitative Comparison}
Fig. \ref{fig:good_example} shows the qualitative comparison of our proposed PLN with 2D-TAN~\cite{zhang2019learning} and LGI~\cite{mun2020local}.
Overall, our proposed PLN performs better on the first three examples. Especially for the first two examples, where the length proportion of target moments in the entire video are relatively short, our proposed model significantly outperforms 2D-TAN and LGI. 
The results to some extent show the advantage of our propped model for handling relatively short target moments (The quantitative verification is shown in Section \ref{ssec:type}).
Additionally, we found that our PLN model generally performs better on videos where the target moment is visually distinguished with the around frames (see the third example) than that on videos where the target moment is visually very similar to the around frames (see the last example).
Among the PLN variants, PLN$_{2,1}$ indicates our proposed two-stage PLN using the predicted score map of stage 1 for prediction, while PLN$_{2,2}$ is the same two-stage model but uses the output of stage 2. 
In general, PLN$_{2,1}$ (the early stage) gives relative coarse predictions, while LN$_{2,2}$ (the later stage) gives relative more accurate predictions.
For example, in the first example, PLN$_{2,1}$ only gives an IoU score of 0.77,  PLN$_{2,2}$ achieves a higher IoU score of 0.96.
It is worth noting that PLN$_{2,2}$ fuses both coarse and fine-grained information of all stages for localization. The results show the importance of the coarse-to-fine information for language-based moment localization.

Additionally, we also illustrate the predicted score maps of PLN$_{2,1}$ and PLN$_{2,2}$ in Fig. \ref{fig:vis}. In a specific score map, each score indicates its predicted relevance of a specific candidate moment with respect to the given query. Besides, candidate moments that are closer to each other in the score map typically have larger IoU between each other. As shown in the predicted score maps of our PLN$_{2,2}$, moments near the target moment give high scores, while moments far from the target moment have relatively low scores. The results show that our model is able to distinguish which moment is more relevant to the given query.

\subsection{Ablation Studies.}\label{ssec:ablation}

In this section, we first investigate the impact of major design choices, \ie ``\textit{Which prediction strategy?}'', ``\textit{How many stages?}'' to better understand our proposed PLN. Additionally, we study the contribution of upsampling connection (UC) and conditional feature manipulation (CFM) in PLN.

\begin{table}[hb!]
\caption{Performance of PLN using different prediction strategies on the TACoS dataset. $t$ denotes which stage is selected for strategy 1. For example, $t$=$1$ means that the predicted relevance score map of the first stage is used for prediction.} 
  \label{tab:pred}
  \centering
  \scalebox{0.8}{%
  \begin{threeparttable}
    \begin{tabular}{@{}l*{8}{c} @{}} 
      \toprule
      \multirow{2}{*}{\textbf{Method}} &
      \multicolumn{3}{c}{\textbf{Rank@1,IoU=m}} &&
      \multicolumn{3}{c}{\textbf{Rank@5,IoU=m}} &
      \multirow{2}{*}{mIoU} \\
      \cmidrule{2-4} \cmidrule{6-8}
      & 0.3   & 0.5   & 0.7   &&  0.3   & 0.5   & 0.7   &       \\ 
      \hline
      \textbf{two-stage:}         &  &  &        &&  &  &      &      \\
      Strategy 1 ($t$=$1$)          & 37.19 & 22.84 & 9.87       && 60.11 & 42.24 & 18.30     & 24.63      \\
      Strategy 1 ($t$=$2$)          & \textbf{43.89} & \textbf{31.12} & \textbf{16.10}  && 65.11 & \textbf{52.89} & \textbf{27.52}  & \textbf{29.70}\\
      Strategy 2                & 43.59 & 30.89 & 15.87       && \textbf{65.28} & 52.54 & 27.44     & 29.43     \\
      \hline
      \textbf{three-stage:}         &  &  &        &&  &  &      &      \\
      Strategy 1 ($t$=$1$)          & 34.29 & 20.72 & 7.62       && 58.36 & 39.49 & 15.30      & 22.75       \\
      Strategy 1 ($t$=$2$)          & 37.97 & 26.37 & 11.52  && 63.13 & 49.34 & 22.72  & 25.79 \\
      Strategy 1 ($t$=$3$)          & \textbf{39.54} & \textbf{28.57} & \textbf{13.35}  && \textbf{64.01} & 49.86 & \textbf{24.12}  & \textbf{26.89} \\
      Strategy 2                    & 39.22 & 28.29 & 12.87  && 63.88 & \textbf{50.24} & 22.94  & 26.54 \\
      \bottomrule
    \end{tabular}
    \end{threeparttable}
  }
\end{table}

\subsubsection{Which Prediction Strategy?}
In this experiment, we use the two-stage and three-stage PLN as our base models to investigate the two different prediction strategies. Table~\ref{tab:pred} summarizes the performance on TACoS. 
For Strategy 1, models using the score map of the later stage achieve better performance. For example, for the three-stage model, its $mIoU$ score gradually improves from 22.75, 25.79, to 26.89 as the stage increases. 
Additionally, in general, Strategy 2 is slightly worse than Strategy 1 which utilizes the prediction of the last stage. Recall that Strategy 2 fuses the score maps of all stages for prediction.
This result is inconsistent with the conclusion that fusion typically gains some performance improvement in other multimedia-related tasks \cite{dong2018cross}.
We attribute it to that the later stage in PLN has already absorbed the information of the previous stage, thus further fusing the results of previous stages has less impact.
Therefore, we use Strategy 1 based on the output of the last stage as our default inference strategy.

\begin{table}[htb!]
\caption{Performance of PLN with the different number of stages on the TACoS dataset. Numbers separated with a hyphen `-' denote the numbers of generated clips used in each stage. Two-stage PLN strikes the best balance between model capacity and generalization ability.  }
\label{tab:stages}
\scalebox{0.8}{
  \centering
  \begin{threeparttable}
    \begin{tabular}{@{}l*{8}{c} @{}} 
      \toprule
      \multirow{2}{*}{\textbf{Method}} &
      \multicolumn{3}{c}{\textbf{Rank@1,IoU=m}} &&
      \multicolumn{3}{c}{\textbf{Rank@5,IoU=m}} &
      \multirow{2}{*}{\textbf{mIoU}} \\
      \cmidrule{2-4} \cmidrule{6-8}
      & 0.3   & 0.5   & 0.7   &&  0.3   & 0.5   & 0.7   &       \\ 
      \hline
      \textbf{one-stage:}         &  &  &        &&  &  &      &      \\
      $N^t$=16                    & 30.93 & 17.22 & 7.03        && 55.52 & 33.54 & 12.31     & 20.24     \\
      $N^t$=32                    & 35.04 & 20.54 & 8.42        && 58.66 & 39.62 & 16.40     & 22.92     \\
      $N^t$=64                    & 37.77 & 22.92 & 10.52       && 60.36 & 45.26 & 20.99     & 24.79     \\
      $N^t$=128                   & 36.32 & 23.82 & 11.42       && 60.96 & 46.59 & 23.64     & 24.51      \\
      \hline
      \textbf{two-stage:}         &  &  &        &&  &  &      &      \\
      $N^t$=32-128                & \textbf{43.89} & \textbf{31.12} & \textbf{16.10}  && \textbf{65.11} & \textbf{52.89} & \textbf{27.52}  & \textbf{29.70}\\
      \hline
      \textbf{three-stage:}         &  &  &        &&  &  &      &      \\
      $N^t$=32-64-128             & 39.54 & 28.57 & 13.35       && 64.01 & 49.86 & 24.12     & 26.89     \\
      \hline
      \textbf{four-stage:}         &  &  &        &&  &  &      &      \\
      $N^t$=16-32-64-128        & 38.42  & 25.17  & 11.65     && 63.63  & 46.96  & 22.99 & 25.44     \\
      \bottomrule
    \end{tabular}%
    \end{threeparttable}
}
\end{table}

\subsubsection{How Many Stages?}

Table \ref{tab:stages} summarizes the performance of PLN variants with the different number of stages on TACoS.
In terms of mIoU, the models with multiple stages consistently beat the one-stage counterparts by a clear margin. The result shows the effectiveness of localizing moments by multiple stages. 
For multi-stage models, the two-stage one turns out to be better than the three-stage and four-stage models. We conjecture this is due to relatively limited number of training video on TACoS, only 75 training vidoes. As the stage of the model increases, the training difficulty of the model also increases, accordingly requiring more data for training. Therefore, we conjecture that the training videos on TACoS are not enough for training the three-stage and four-stage models. 
On ActivityNet Captions and Charades-STA,  the three-stage and four-stage models are comparable to the two-stage model (See the supplementary materials). Considering the two-stage PLN achieves the best overall performance, we use the two-stage PLN as the default setting.

\begin{table}[htb!]
  \caption{Effectiveness of the coarse-to-fine manner on TACoS. * denotes the coarse-to-fine manner. }
  \label{tab:coarse2fine}
  \centering
 \scalebox{0.8}{%
 \begin{threeparttable}
    \begin{tabular}{@{}l*{8}{c} @{}}
     \toprule
      \multirow{2}{*}{\textbf{Method}} &
      \multicolumn{3}{c}{\textbf{Rank@1,IoU=m}} &&
      \multicolumn{3}{c}{\textbf{Rank@5,IoU=m}} &
      \multirow{2}{*}{\textbf{mIoU}} \\
      \cmidrule{2-4} \cmidrule{6-8}
      & 0.3   & 0.5   & 0.7   &&  0.3   & 0.5   & 0.7   &       \\
      \midrule
      $N^t$=64                    & \textbf{37.77} & 22.92 & 10.52       && 60.36 & 45.26 & 20.99     & 24.79     \\
      $N^t$=32-32                 & 34.54 & 20.62 & 7.72        && 59.94 & 41.31 & 17.15     & 22.63      \\
      $N^t$=64-64                 & 35.67 & 21.07 & 9.20        && 59.29 & 43.64 & 20.64     & 23.00     \\
      $N^t$=32-64*                & 37.29 & \textbf{23.69} & \textbf{11.12} && \textbf{61.93} & \textbf{45.69} & \textbf{21.19} & \textbf{25.01} \\
      \midrule
      $N^t$=128                   & 36.32 & 23.82 & 11.42       && 60.96 & 46.59 & 23.64     & 24.51      \\
      $N^t$=32-32                 & 34.54 & 20.62 & 7.72        && 59.94 & 41.31 & 17.15     & 22.63 \\
      $N^t$=128-128               & 33.34 & 19.65 & 9.55        && 58.69 & 43.44 & 22.57     & 22.22     \\
      $N^t$=32-128*               & \textbf{43.89} & \textbf{31.12} & \textbf{16.10} && \textbf{65.11} & \textbf{52.89} & \textbf{27.52} & \textbf{29.70} \\
      \hline
    \end{tabular}%
    
    \end{threeparttable}
  }

\end{table}

\subsubsection{Coarse-to-Fine Manner?}
To verify the viability of our devised coarse-to-fine manner for PLN, we compare it with the PLN using the same temporal granularity of video clips in all stages ($N^t$ is the same for all stages).
As shown in Table \ref{tab:coarse2fine}, our PLN with the coarse-to-fine temporal granularities (line 4 in each table block) outperforms the model with the same temporal granularity across stages (line 2, 3), showing the superiority of the coarse-to-fine manner for the multi-stage PLN.
Additionally, we also report the results of one-stage model (line 1), and find it better than the two-stage model without the coarse-to-fine temporal granularities (line 3).
The result reveals the necessity of the coarse-to-fine manner for multi-stage localization.


\begin{table}[tb!]
\caption{Ablation Studies Over the Two-stage PLN. Our Full Model Performs the Best on three datasets.}
  \label{tab:ablation}
  \centering
\scalebox{0.8}{%
\begin{threeparttable}
    \begin{tabular}{@{}ll*{6}{c} @{}} 
     \toprule
      \multirow{2}{*}{\textbf{Dataset}} &
      \multirow{2}{*}{\textbf{Method}} &
      \multicolumn{3}{c}{\textbf{Rank@1,IoU=m}} &
      \multirow{2}{*}{mIoU} \\
      \cmidrule{3-5} 
      && 0.3   & 0.5   & 0.7   &         \\ 
       \midrule
      \multirow{3}{*}{\textbf{Charades-STA}}
      & Full model          & \textbf{59.33} & \textbf{45.43} & \textbf{26.26}  & \textbf{41.28}\\
      & w/o CFM             & 57.50 & 44.09 & 25.89       &  40.19     \\
      & w/o UC              & 56.10 & 43.52 & 26.40       &  39.23     \\
     \midrule
       \multirow{3}{*}{\textbf{TACoS}}
      & Full model          & \textbf{43.89} & \textbf{31.12} & \textbf{16.10}  & \textbf{29.70}\\
      & w/o CFM             & 43.01 & 30.97 & 15.22       &  29.05     \\
      & w/o UC              & 36.09 & 22.44 & 10.77       &  23.82     \\
      \midrule
      
    \multirow{3}{*}{\textbf{ActivityNet Captions}}
      & Full model          & \textbf{59.65} & \textbf{45.66} & \textbf{29.28}  & \textbf{44.12}\\
      & w/o CFM             & 58.99 & 44.61 & 28.44       &  43.71     \\
      & w/o UC              & 59.19 & 44.52 & 27.97       &  43.39     \\
     
      \bottomrule
    \end{tabular}%
    \end{threeparttable}
 }
  
\end{table}

\subsubsection{The Effectiveness of UC and CFM}
In order to investigate the contribution of upsampling connection (UC) and conditional feature manipulation (CFM) in the proposed PLN, we conduct ablation studies and summarize the results in Table \ref{tab:ablation}.
On Charades-STA, we observe that our full model performs the best in terms of all metrics.
Removing each component from PLN, \ie UC or CFM, would result in performance degeneration. The result demonstrates the effectiveness of each component in our PLN.

\subsection{Analysis on Different Types of Moments}\label{ssec:type}

\begin{figure}[tb!]
\centering
\subfigure[]{
\label{fig:tacos_gf}
\includegraphics[width=0.47\linewidth]{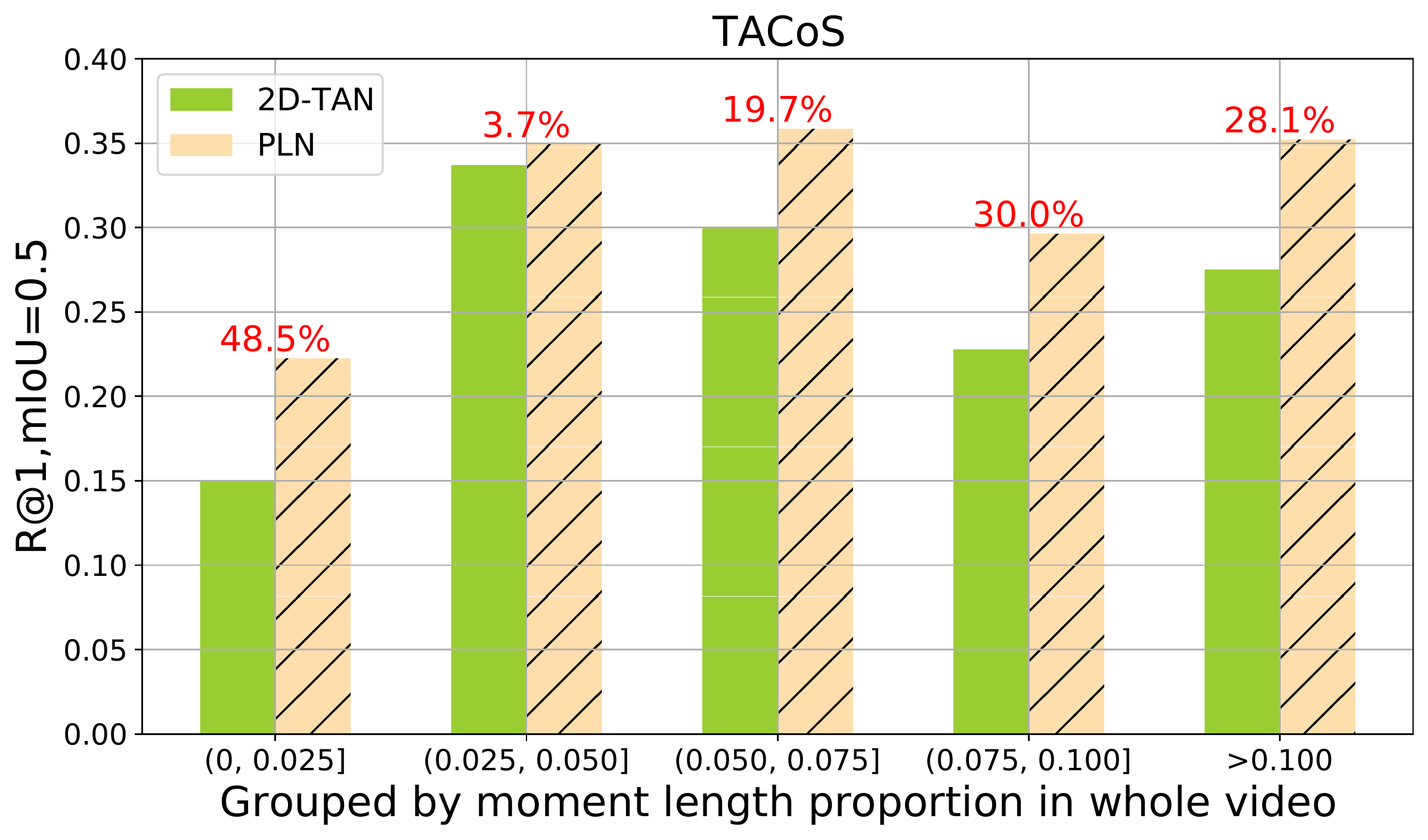}
}
\quad
\subfigure[]{
\includegraphics[width=0.47\linewidth]{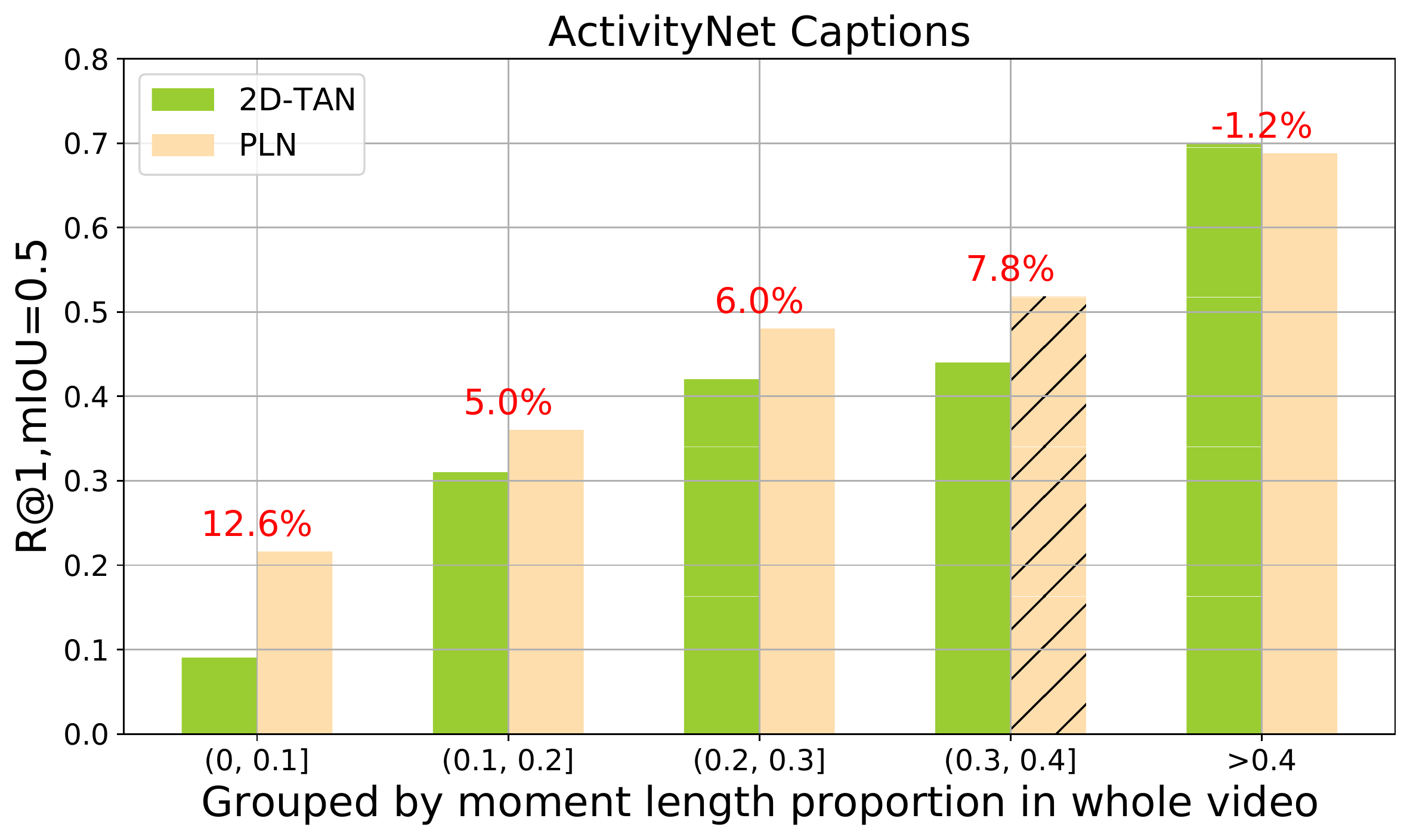}
\label{fig:ac_gf}
}
\caption{Detailed performance comparison between 2D-TAN and our proposed PLN on (a) TACoS  and (b) ActivityNet Captions. Moments have been grouped in terms of their length proportion in the entire video. The red number over the bins is the relative performance improvement of PLN over 2D-TAN.}
  \label{fig:group_per}
\end{figure}

To investigate how our multi-stage PLN performs on different groups of moments, we group 
target moments according to their length proportion of the entire video length in the test set. We compare  PLN with the 2D-TAN model that to some extent can be seemed as a one-stage PLN.
As shown in Fig.~\ref{fig:group_per}, we observe a phenomenon that the both models perform the worst on the first group where moments have the shortest average temporal duration.
It to some extent demonstrates that localizing the short moments is more challenging.
In almost groups, our PLN consistently outperforms 2D-TAN, which again indicates the effectiveness of our PLN for language-based moment localization.
Especially, PLN achieves the highest relative performance improvement of 48.5\% in the first group on TACoS and 12.6\% on ActivityNet Captions. It demonstrates that our proposed multi-stage localization mechanism with the coarse-to-fine manner can better handle the relatively short target moments in long videos than one-stage 2D-TAN.

\section{Conclusion}

This paper shows the viability of resolving language-based moment localization in a progressive coarse-to-fine manner.
We contribute a novel multi-stage Progressive Localization Network which is capable of localizing the target moment progressively via multiple localization branches. The localization branches are connected via a conditional feature manipulation module and an upsampling connection, making the later stage absorb the previously learned coarse information, thus facilitate the more fine-grained localization.
Besides, we also show the potential of our multi-stage PLN for localizing short moments in long videos, which has been ignored by previous works. We believe that this simple and effective multi-stage progressive architecture can be of interest to many language-based moment localization research efforts. 
In future work, we will apply the idea of coarse-to-fine localization to other related tasks such as video object grounding~\cite{yang2020weakly,xiao2020visual}, and video relation detection ~\cite{li2021interventional,shang2019annotating}.

\begin{acks}
This work was supported by National Key R\&D Program of China (No. 2018YFB1404102), NSFC (No. 61902347, 61976188, 62002323), the Public Welfare Technology Research Project of Zhejiang Province (No. LGF21F020010), the Research Program of Zhejiang Lab (No. 2019KD0AC02), the Fundamental Research Funds for the Provincial Universities of Zhejiang.

\end{acks}

\bibliographystyle{ACM-Reference-Format}
\bibliography{reference}

\end{document}